\documentclass[runningheads]{llncs}

\usepackage{eccv}

\usepackage{eccvabbrv}

\usepackage{graphicx}
\usepackage{booktabs}

\usepackage[accsupp]{axessibility}  

\usepackage[marginal]{footmisc}
\renewcommand{\thefootnote}{}

\usepackage{hyperref}

\usepackage{orcidlink}

\usepackage{multirow}
\usepackage{booktabs}

\usepackage{makecell}
\usepackage{changes}
\usepackage{arydshln}
\usepackage{graphicx}

\usepackage{colortbl}%

\definecolor{mygray}{gray}{.9}

\usepackage{marvosym}

\begin{document}

\title{AdaDiffSR: Adaptive Region-aware Dynamic Acceleration Diffusion Model for\\ Real-World Image Super-Resolution} 

\titlerunning{AdaDiffSR: Adaptive Dynamic Acceleration Diffusion Model for Real SR}

\author{Yuanting~Fan\inst{1,2}\textsuperscript{*}\orcidlink{0009-0008-6507-666X} \and
Chengxu~Liu\inst{2,3}\textsuperscript{*}\orcidlink{0000-0001-8023-9465} \and
Nengzhong~Yin\inst{2} \and
Changlong~Gao\inst{2} \and
Xueming~Qian\inst{2,3}\textsuperscript{\Letter}\orcidlink{0000-0002-3173-6307}
}
\authorrunning{Y. Fan et al.}

\institute{School of Software Engineering, Xi’an Jiaotong University, Xi'an, China \and
Xi’an Jiaotong University, Xi'an, China \and
Shaanxi Yulan Jiuzhou Intelligent Optoelectronic Tech. Co., Ltd, Xi'an, China
\email{\{retofan,chengxuliu,ynz0608,gaochanglong\}@stu.xjtu.edu.cn, qianxm@mail.xjtu.edu.cn}}

\maketitle

\renewcommand{\thefootnote}{\fnsymbol{footnote}} 
\footnotetext[1]{Equal contribution.}
\begin{abstract}
\label{sec:abstract}
    Diffusion models (DMs) have shown promising results on single-image super-resolution and other image-to-image translation tasks. Benefiting from more computational resources and longer inference times, they are able to yield more realistic images. 
    Existing DMs-based super-resolution methods try to achieve an overall average recovery over all regions via iterative refinement, ignoring the consideration that different input image regions require different timesteps to reconstruct.
    In this work, we notice that previous DMs-based super-resolution methods suffer from wasting computational resources to reconstruct invisible details. 
    To further improve the utilization of computational resources, we propose AdaDiffSR, a DMs-based SR pipeline with dynamic timesteps sampling strategy (DTSS). Specifically, by introducing the multi-metrics latent entropy module (MMLE), we can achieve dynamic perception of the latent spatial information gain during the denoising process, thereby guiding the dynamic selection of the timesteps. 
    In addition, we adopt a progressive feature injection module (PFJ), which dynamically injects the original image features into the denoising process based on the current information gain, so as to generate images with both fidelity and realism.
    Experiments show that our AdaDiffSR achieves comparable performance over current state-of-the-art DMs-based SR methods while consuming less computational resources and inference time on both synthetic and real-world datasets.
\end{abstract}

\section{Introduction}
\label{sec:intro}

Real-world image super-resolution aims to recover a realistic high-resolution (HR) image from an unknown degraded low-resolution (LR) counterpart. It is adopted to enhance the image visual quality and is widely used in the fields of satellite imagery~\cite{liebel2016single}, surveillance, etc. 
Nowadays, significant advances have been made in diffusion models for image synthesis tasks~\cite{jimenez2023mixture, mokady2023null,wu2023tune, hertz2022prompt}. 
Recent research demonstrates that the pre-trained models~(\eg, SD~\cite{rombach2022high}) can easily transferred to various downstream low-level vision tasks, including image and video translation~\cite{rombach2022high, sahak2023denoising, saharia2022image, wang2023exploiting}. 
In this work, we explore the potential of using pre-trained diffusion models to perform super-resolution efficiently and effectively.

\begin{figure}[t]
  \centering
   \includegraphics[width=\linewidth]{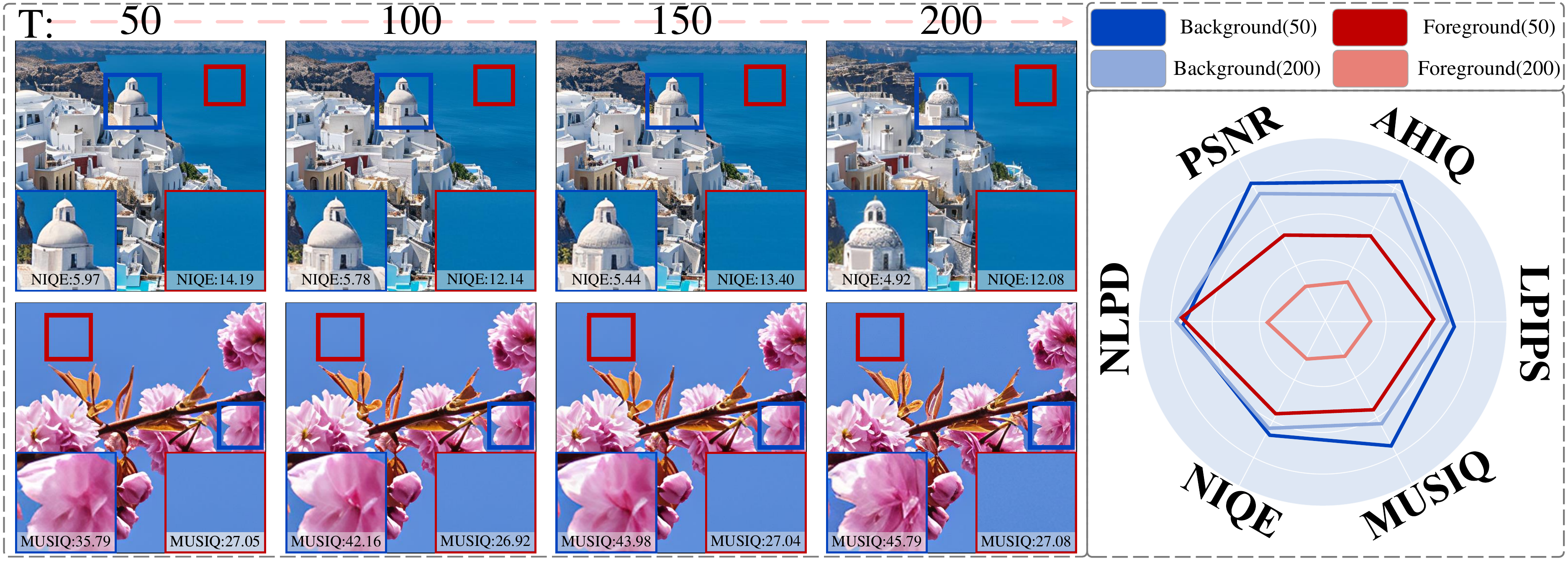}
   \caption{Visual comparisons between background and foreground regions during denoising process. The red and blue boxes represent the background and foreground regions, respectively. We visualize the variations of several corresponding metrics as the radar chart on the right, the smaller the better. As the timesteps increase from 50 to 200, we find that the visual results of the foreground regions become more satisfactory while the background remains almost unchanged.}
   \label{fig:teaser}
\end{figure}

According to whether using pre-trained parameters, Existing DMs-based SR methods~\cite{rombach2022high, sahak2023denoising, saharia2022image, wang2023exploiting} can be divided into two paradigms. The former~\cite{saharia2022image,sahak2023denoising} trains the DMs-based SR model from scratch while using LR features as additional input to constrain the fidelity of output. 
The latter~\cite{wang2023exploiting,rombach2022high} incorporates constraints into the denoising progress of the pre-trained diffusion models, which not only avoids the demand for computational resources to train the diffusion models, but also enables better utilization of the diffusion prior within the pre-trained diffusion models. 
However, all of the above methods inherently fail to take into account the relationship between reconstructing different image regions and the computational resources they cost. It means that some regions may require fewer timesteps to achieve a satisfactory reconstruction performance, while others require larger timesteps. 
Consequently, these methods exhibit sub-optimal utilization of computational resources. As shown in Fig.~\ref{fig:teaser}, with the increasing of timesteps in the denoising process, the background regions~(\eg, sky or sea in red boxes) are not visually distinct, and only the foreground regions~(\eg, surfaces or structures in blue boxes) become more realistic.

In this paper, we propose a novel DMs-based SR pipeline, dubbed AdaDiffSR. It utilizes the latent information gain to dynamically adjust timesteps during inference and effectively reduces the computational resource overhead. 
Specifically, we divide the entire input image into a series of sub-regions separately for the denoising process. For these sub-regions, as shown in Fig.~\ref{fig:network}, we design a dynamic timesteps sampling strategy~(DTSS) that adjusts timesteps dynamically according to the information gain within these sub-regions. To implement it, we propose a multi-metrics latent entropy~(MMLE) module, which incorporates multiple image quality assessment~(IQA) metrics and can perceive information gain dynamically from both qualitative and quantitative perspectives. 
In addition, to balance the fidelity of the reconstructed image with the texture generation capability of diffusion models, we further propose a progressive feature injection~(PFJ) module. It dynamically adjusts the fusion process between original input images and denoising features according to the current information gain. Meanwhile, by applying the latent space region integration strategy, discontinuities between different restored sub-regions are effectively eliminated.

Our main contributions are as follows:
\begin{itemize}
    \item We introduce AdaDiffSR, a diffusion model for real-world image super-resolution. It is one of the first works to focus on optimizing the utilization of computational resources in DMs-based SR paradigms and provides a new inspiration for future research on denoising models. 
    \item We design a dynamic timesteps sampling strategy (DTSS), in conjunction with the proposed multi-metrics latent entropy~(MMLE) module, which can significantly reduce the computational overhead of the denoising process while obtaining superior results. 
    \item We propose a progressive feature injection~(PFJ) module to dynamically incorporate original image features into the denoising process, allowing the recovered results with fidelity and rich texture details. 
    \item Extensive experiments demonstrate that the proposed AdaDiffSR can significantly reduce computational resources and inference time compared to existing SOTA methods. 
\end{itemize}

\section{Related Work}
\label{sec:related}

\subsection{Single Image Super-Resolution}
Single Image Super-Resolution~(SISR) aims to restore the corresponding HR counterpart from the degraded LR image. The previous SR methods~\cite{dong2015image,dong2016accelerating,Lim_2017_CVPR_Workshops,ledig2017photo,liebel2016single} focused on learning the mappings between LR and HR image for particular degradation kernel~(\eg, Bicubic downsampling or Gaussian blur kernel). However, the above mappings have limited generalization capacity in real-world scenarios and cannot achieve better visual performance. 

Recently, more works~\cite{fritsche2019frequency,maeda2020unpaired,wang2021unsupervised,zhang2021blind,liu2024motion,liu2022learning,liu2024decoupling,wu2024seesr,chen2024iterative} have focused on optimizing the performance of real-world SR~(or blind SR), which aims to learn the similar degradation process in real-world scenarios. Due to the lack of paired real-world training datasets, some methods~\cite{zhang2021designing,wang2021real} explicitly synthesize the paired LR-HR images for blind SR. 
Specifically, recent works~\cite{zhang2021designing, wang2021real} present the effective random degradation process, which produces the LR counterparts by applying blur before downsampling the HR images, and then adding noise and applying JPEG compression to the downsampled results. These degradation schemes have been crucial for GAN-based SR methods to achieve state-of-the-art performance.
Building upon these schemes, recent DMs-based SR methods~\cite{sahak2023denoising,saharia2022image,wang2023exploiting, yue2024resshift} further show more satisfactory performance in real-world scenarios. 
However, the above SR methods failed to consider the utilization of computational resources during the reconstruction process.
In this work, we consider the latent information gain and adjust timesteps of different image regions dynamically during the denoising process, thus reducing the computational resource overhead effectively.

\subsection{Diffusion Model in Super-Resolution}
With the rapid development of diffusion models~(DMs) in the image and video generation tasks~\cite{wu2023tune}, recently, numerous DMs-based SR methods~\cite{sahak2023denoising,saharia2022image,wang2023exploiting, yue2024resshift,mei2024codi,lin2023diffbir} have been proposed. 
Compared to the previous GAN-based SR methods~\cite{zhang2021blind, wang2018esrgan, wang2021real} suffer from mode collapse~\cite{che2016mode, srivastava2017veegan, mao2019mode} and convergence difficulty during the training process, the existing DMs-based SR methods obtain the SR results via iteratively refine the original noise images, which significantly enhances the model robustness and generalization capacity. 
Specifically, SR3~\cite{saharia2022image} first introduces the diffusion models into the SR tasks and achieves state-of-the-art performance on the face and natural datasets. Recent works~\cite{sahak2023denoising, wang2023exploiting} have attempted to improve DMs-based SR methods from the perspectives of image degradation and image fidelity. 
In this work, we consider the utilization of computational resources in DMs-based SR methods and propose to dynamically adjust the timesteps of different image regions during the denoising process.

\subsection{Adaptive Inference}
The previous methods~\cite{yu2019autoslim,yu2019universally,yu2018slimmable} focus more on designing slimmable network structures, which adaptively approximate the network performance according to the data characteristics. 
Specifically, AutoSlim~\cite{yu2019autoslim} trains a single slimmable network to adjust the network performance of different channel configurations, and then searches the optimized channel configurations under different resources. 

Instead of adjusting the network structure, inspired by the fact that different regions have different restoration difficulties, ClassSR~\cite{kong2021classsr} utilizes the characteristics of different image regions by dividing the images into patches, then apply different capacity models to the corresponding restoration difficulties. However, the application of multiple models leads to a significant increase in model parameters.
APE~\cite{wang2022adaptive} proposes to learn the incremental capacity of each model layer instead of restoration difficulties of patches, enabling the patch to exit at the optimal layer, thus achieving more practical speedup. 
Although existing methods~\cite{wang2021exploring,liu2020deep,verelst2020dynamic} have taken into account the information variation~(network layer-wise) of different image regions during the reconstruction process, they lack further exploration of the latent information variation~(timestep-wise) during the denoising process in diffusion models. 
In this work, we combine multiple IQA metrics to measure the latent information gain during the denoising process in DMs-based SR methods, thus guiding the dynamic timesteps selection and reducing the computational resource overhead effectively.

\begin{figure}[t]
  \centering
   \includegraphics[width=\linewidth]{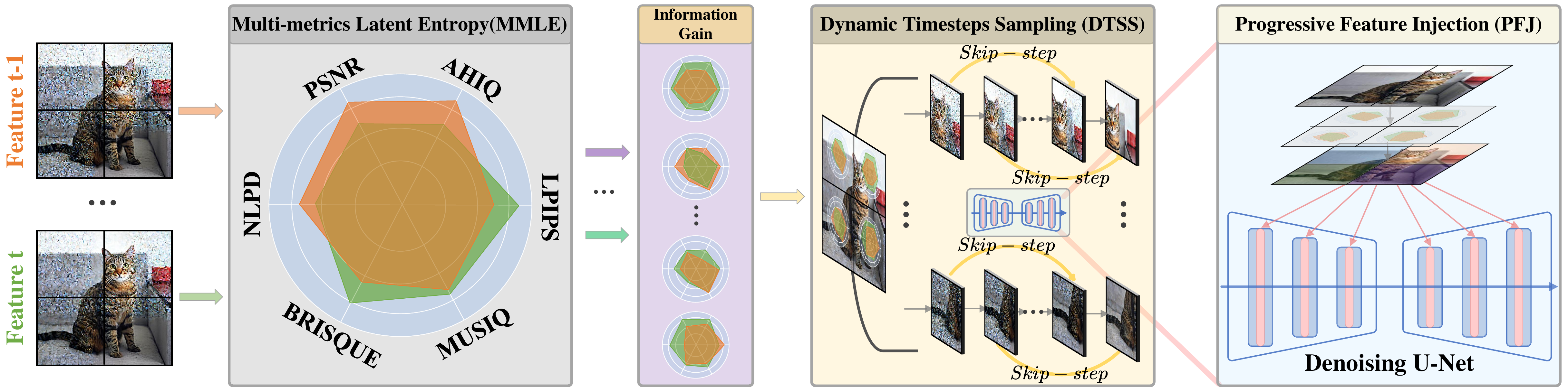}
   \caption{The framework of AdaDiffSR. We calculate the information gain during the denoising process, via information gain, we can modulate the original image features to guide the PFJ module and adjust timesteps dynamically for different regions, thus achieving a trade-off between the computational resources and restoration quality.}
   \label{fig:network}
\end{figure}

\begin{figure}[t]
  \centering
   \includegraphics[width=\linewidth]{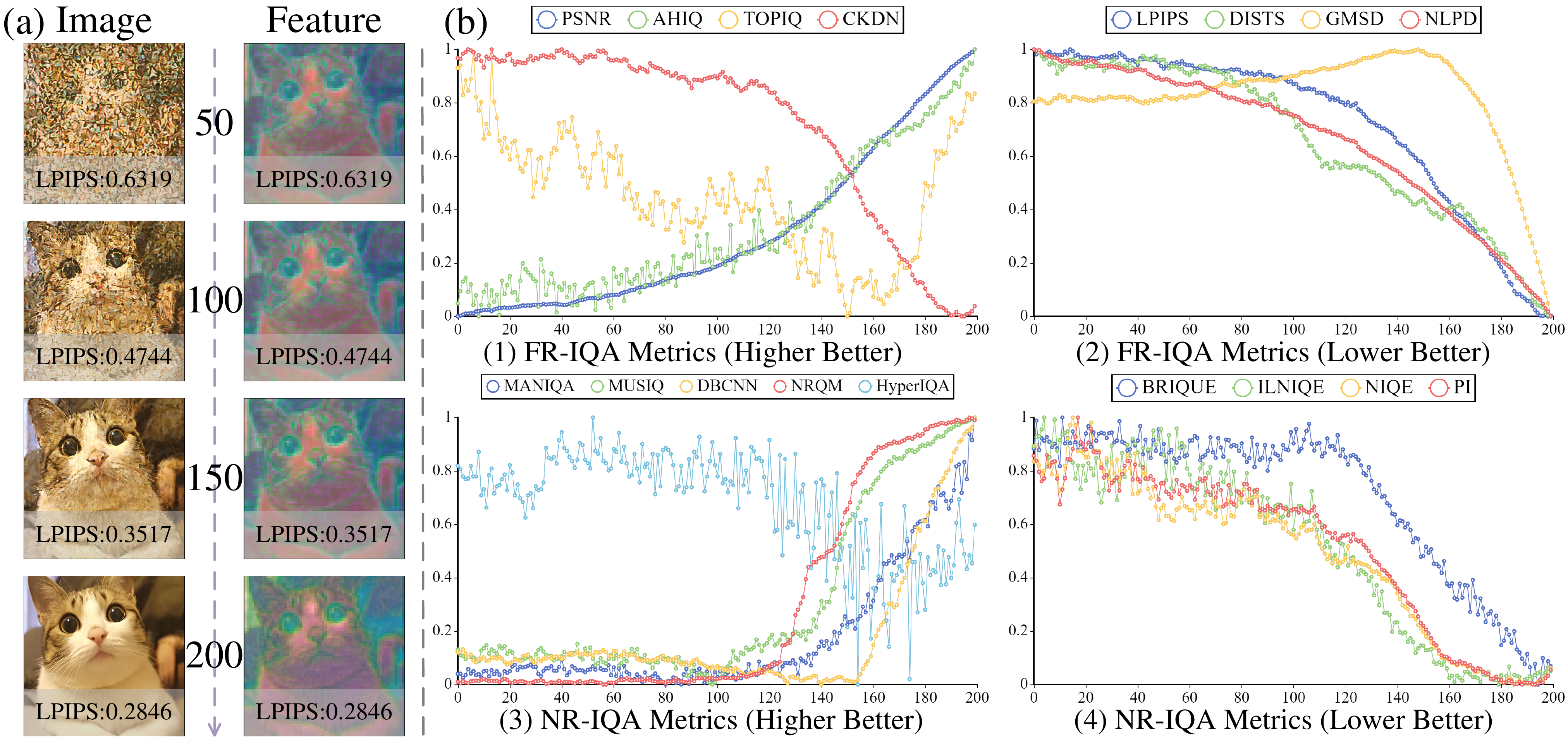}
   \caption{To demonstrate the validity of different IQA metrics during the denoising process. (a), we visualize the feature in perceptual-oriented metric LPIPS and the corresponding denoised image during the restoration process. (b), we plot the variations of several IQA metrics during the denoising process with 200 timesteps using DDPM sampling strategy~\cite{ho2020denoising}. Best viewed when disabled some metrics~(\eg, TOPIQ~\cite{chen2023topiq}, CLIP-IQA~\cite{wang2022exploring} and etc.), the variations of these metrics are similar with HyperIQA~\cite{su2020blindly}.}
   \label{fig:mmle}
\end{figure}

\section{Methodology}
\label{sec:method}

\subsection{Motivation}
\label{subsec:motivation}
To reduce the computational resource overhead of DMs-based SR methods, we observe that although latent diffusion models can significantly reduce computational resource overhead by reducing feature size, as shown in Fig.~\ref{fig:teaser}, there are still a lot of computational resources used to reconstruct invisible details, which leads to sub-optimal utilization of computational resources. 
Therefore, we propose AdaDiffSR to further improve the utilization of computational resources. 
As shown in Fig.~\ref{fig:network}, AdaDiffSR consists of the following three components: multi-metrics latent entropy module~(MMLE) as shown in Sec.~\ref{subsec:mmle}, dynamic timesteps sampling strategy~(DTSS) as shown in Sec.~\ref{subsec:dts}, and the progressive feature injection module~(PFJ) and region integration strategy are as shown in Sec.~\ref{subsec:tradeoff}.

\subsection{Multi-metrics latent entropy module}
\label{subsec:mmle}
Based on the previous IQA metrics~\cite{lao2022attentions,chen2023topiq,zheng2021learning,zhang2018unreasonable,ding2020image,xue2013gradient,yang2022maniqa,ke2021musiq,zhang2020blind,ma2017learning,golestaneh2021no,wang2022exploring,ying2020patches,su2020blindly,kang2014convolutional,mittal2012no,zhang2015feature,mittal2012making,blau20182018}, we propose the multi-metrics latent entropy module~(MMLE) to measure the information evolution from many different perspectives during denoising process. 
More precisely, we calculate these metrics in latent space, and we find that many perceptual-oriented IQA metrics fail to reflect the information evolution process accurately. 
On the contrary, the traditional psnr-oriented metrics accurately perceive the information evolution. 
The above phenomenon is due to the massive Gaussian noise incorporated during the denoising process. 

As shown in Fig.~\ref{fig:mmle}(b), we plot the variations of IQA metrics during the denoising process. To make it easier to observe the trends in variations of these metrics, we divide these metrics into two categories~(\ie, NR-IQA metrics~(No-Reference) and FR-IQA~(Full-Reference) metrics) and normalize all metrics between zero and one.
Significant differences between different IQA metrics during the denoising process allow us to divide them into three categories of variations. 
Specifically, the first category of variations positively correlates with the information evolution process, including PSNR, AHIQ, etc. 
The second category of variations has no significant correlation with the information evolution process, mainly including TOPIQ, CKDN, HyperIQA, etc. 
The remaining variations are categorized into the third category, which tends to significantly reflect the information evolution in the second half of the denoising process. 
Since most of the third category of metrics utilize CNN or Transformer~\cite{vaswani2017attention} to extract features in latent space, and in the first half of the denoising process, the latent features are not clear enough to reflect image quality due to excessive Gaussian noise. As shown in Fig.~\ref{fig:mmle}(a), the noise is gradually eliminated, and the latent feature tends to be apparent, allowing these metrics to perceive the information evolution.

Based on the above conclusions, we select several IQA metrics with different perspectives for discriminating the multi-dimensional information gain between two latent features, thus guiding the subsequent dynamic timesteps sampling strategy~(DTSS) and progressive feature injection~(PFJ) module. 
Specifically, we define the information gain $I_{i}$, which measures the latent representation performance between current timestep $i$ and the previous timestep $i-1$:
\begin{equation}
I_{i} = \sigma (R_{i} - R_{i-1}),
\label{eq:infogain}
\end{equation}
where $\sigma(\cdot)$ represents the tanh function, $R_{i}$ is the latent representation performance of timestep $i$. 
Since information gain stems from multi-dimensional metrics, evolution in some dimensions may lead to degradation in other dimensions, through tanh function, we limit the range of $I_{i}$ to $[-1,1]$ to perceive this degradation. 
As shown in Fig.~\ref{fig:mmle}(b), FR-IQA metrics are more robust to the information evolution, which is mainly reflected in the gradient continuity of the corresponding curve. 
On the contrary, NR-IQA metrics suffer from the limitation of without reference image~(\ie, original image feature) and cannot perceive the image quality improvement until the second half of the denoising process. 
To achieve the trade-off between capacity and efficiency, we select four FR-IQA metrics~(\ie, PSNR, LPIPS, AHIQ, and NLPD) and two NR-IQA metrics~(\ie, BRISQUE and MUSIQ) to measure the information gain:
\begin{equation}
\begin{split}
R_{i} = \sum_{c \in C} {\omega_{c} \times  M_{c}(f_{i}, o)},
\end{split}
\label{eq:representation}
\end{equation}
where $f_{i}$ and $o$ represent the denoising feature in timestep $i$ and the original image feature, respectively. $\omega_{c}$ denotes the coefficients of different IQA metrics, which are used to balance the influence of different metrics.
$M_{c}$ means the normalized value of the corresponding IQA metric. 
$C$ consists of the six metrics mentioned above. 
Note that for the robustness of information evolution, we only calculate the two NR-IQA metrics information gain during the second half of the denoising process. 
During inference, considering the tremendous time cost of calculating IQA metrics on the fly, we propose to train a lightweight regressor that consists of multiple convolutional layers and global average pooling layers to estimate the multi-dimensional information gain. 
More analysis of the IQA metrics selection and the structure of regressor can be found in the ablation studies and supplementary materials. 

\subsection{Dynamic timesteps sampling strategy}
\label{subsec:dts}
Inspired by previous works~\cite{wang2022adaptive, kong2021classsr}, as shown in Fig.~\ref{fig:teaser}, we find that different regions require different computational resources to achieve a trade-off between utilization of computational resources and reconstruction performance. 
Benefiting from the above MMLE regressor, we propose a dynamic timesteps sampling strategy~(DTSS), which improves the utilization of computational resources and adaptively accelerates the DMs-based SR methods.

Specifically, to preserve the prior encapsulated within pre-trained diffusion models, we first crop the input images into multiple sub-regions with the fixed spatial resolution as pre-trained diffusion models. 
Then, we calculate the multi-dimensional information gain using the MMLE regressor during the denoising process, thereby guiding the dynamic timesteps selection for each sub-region.

The specific strategy for dynamic timestep selection is as follows. Firstly, we define the information gain threshold and max timesteps of diffusion models as $\tau$ and $T_{max}$, respectively. 
Then, we generate a codebook to store the sampling-related parameters for skipping from one timestep to another. With this codebook, we can implement the skip-step strategy, which means that we perform multiple original DDPM~\cite{ho2020denoising} or DDIM~\cite{song2020denoising} timesteps at once. 
To improve the inference time and stability of the denoising process, we define four intervals for the skip-step strategy to adjust timesteps dynamically for different regions. 
The conclusion obtained from Sec.~\ref{subsec:mmle} indicates that the FR-IQA metrics are robust, while the NR-IQA metrics depend on the image region context. 
Therefore, we divide these sub-regions into three categories from FR and NR perspectives: 
\begin{itemize}
    \item Stable regions: The FR metrics increasing steadily while the NR metrics are almost unchanged, indicating that the information within these regions is not enhanced in perceptual perspectives~(\eg, the background regions in Fig.~\ref{fig:teaser}). In these regions, we apply larger intervals for the better utilization of computational resources and inference time. 
    \item Growing regions: The FR and NR metrics increasing steadily together~(\eg, the foreground regions in Fig.~\ref{fig:teaser}), the information within these regions is growing along with the denoising process. Then we apply a smaller interval to enhance the reconstruction quality. 
    \item Saturated regions: We also notice the phenomenon that NR metrics significantly declined with more denoising iterations, which is due to the information within these regions being saturated. Therefore we save the best results of these regions in NR perspective and exit the denoising process in advance. 
\end{itemize}

Through the information gain threshold $\tau$, we are able to distinguish between different sub-region categories. 
Note that the category of sub-regions may change during the denoising process, so the interval of skip-step can be modified dynamically. Finally, when the timesteps reach the value of $T_{max}$, the reconstruction is finished and the final equivalent timesteps are much smaller than the $T_{max}$, thus achieving adaptive acceleration while keeping restoration quality. More analysis about interval settings can be found in the ablation studies.

\subsection{Trade-off between fidelity and realism}
\label{subsec:tradeoff}
Benefiting from the powerful generative capacity of diffusion models, the reconstructed images tend to be more realistic. Nevertheless, similar to other generative model based SR methods, the results may lack similarity from the original input~(\ie, lack of fidelity). 
Therefore, we propose the progressive feature injection module~(PFJ) to achieve the trade-off between fidelity and realism. The PFJ module dynamically incorporates latent features of original input images into the denoising process based on the current information gain, thus enhancing fidelity while maintaining realism. 
Specifically, the PFJ module dynamically regulates the fusion intensity of the original image features according to the information gain from different perspectives:
\begin{equation}
\begin{split}
\hat{o} = \alpha \times o + \beta; \ \ \ \ \alpha, \beta = \phi (o,I_{i}),
\end{split}
\label{eq:pfj}
\end{equation}
where $\alpha$ and $\beta$ represent the modulation coefficients of the original image features $o$. And we design a small CNN network $\phi(\cdot)$ to estimate the modulation coefficients from the current timestep information gain $I_{i}$ and original image features $o$. 
Generally, if the NR perspectives information gain is more prominent, which means that the realism of the current timestep denoising feature is increasing. To balance the fidelity and realism, the $\phi(\cdot)$ needs to predict larger modulation coefficients to adjust the weights of the original image features for better fidelity, and vice versa.

Although the problem of image fidelity has been solved, the fixed input resolution of the pre-trained diffusion model limits the further application of DMs-based SR methods. 
The previous CNN-based SR methods used the naive strategy to integrate the overlapping regions, leading to discontinuities in the boundaries. 
Inspired by ~\cite{wang2023exploiting,jimenez2023mixture}, we integrate these regions in latent space using the Gaussian weight maps that are generated from Gaussian kernel, thus avoiding the discontinuities in pixel space, as shown in Fig.~\ref{fig:integration}. 
To speed up the reconstruction while eliminating the mutual noise influence in adjacent regions, we only employ the above integration strategy in the final timestep. 
The details of the region integration strategy can be found in the supplementary materials.

\section{Experiments}
\label{sec:exp}

\subsection{Implementation Details}
\label{subsec:imp}

Following previous work~\cite{wang2023exploiting}, we finetune the pre-trained diffusion model~(Stable Diffusion 2.1-base~\cite{rombach2022high}) using the synthetic training dataset. We follow Stable Diffusion to use Adam optimizer and set the learning rate as $5\times {10}^{-5}$. To avoid affecting the prior within the pre-trained diffusion model, we conduct the fine-tuning process on $512 \times 512$ resolutions. 
During inference, we set the max timesteps $T_{max}$ and information gain threshold as $1,000$ and $5\times 10^{-3}$, respectively. 
To handle the images with arbitrary resolutions, we adopt the region integration strategy mentioned in Sec.~\ref{subsec:tradeoff}. 
To train the MMLE regressor, we follow the Real-ESRGAN~\cite{wang2021real} degradation pipeline to generate synthetic LR-HR pairs with $512 \times 512$ resolutions. Then, we freeze the parameters of the finetuned diffusion model to train the regressor using the $L2$ loss function.

\begin{table*}[t]
        \caption{Quantitative comparisons with the state-of-the-art methods on both synthetic and real-world datasets. The best and second-best performances are in \textcolor{red}{red} and \textcolor{blue}{\underline{blue}} color, respectively. Real-ESRGAN$+$~\cite{wang2021real} is abbreviated as R-ESRG.}
	\label{tab:metrics}
	\centering
        \resizebox{\textwidth}{!}{
	\begin{tabular}{llccccccccc}
            \toprule
		  \multirow{2}{*}{Datasets} & \multirow{2}{*}{Metrics} & \multirow{2}{*}{\makecell{\ RealSR\  \\ \cite{ji2020real}}} & \multirow{2}{*}{\makecell{\ DASR\ \\ \cite{liang2022efficient}}}  & \multirow{2}{*}{\makecell{BSRGAN \\ \cite{zhang2021blind}}}  & \multirow{2}{*}{\makecell{R-ESRG\\ \cite{wang2021real}}}  & \multirow{2}{*}{\makecell{FeMaSR \\ \cite{chen2022real}}}  & \multirow{2}{*}{\makecell{LDM \\ \cite{rombach2022high}}} & \multirow{2}{*}{\makecell{StableSR \\ \cite{wang2023exploiting}}} & \multirow{2}{*}{\makecell{ResShift \\ \cite{yue2024resshift}}} & \multirow{2}{*}{Ours} \\
                    & & & & & & & & & \\
            \midrule
		\multirow{5}{*}{\makecell{DIV2K \\ Valid~\cite{timofte2017ntire}}} & PSNR~$\uparrow$  & 25.11  & \textcolor{red}{28.33} & \textcolor{blue}{\underline{25.26}} & 24.84 & 22.97 & 20.58 & 23.83 & 24.53 & 24.25 \\
                                      & SSIM~$\uparrow$     & 0.7170 & \textcolor{red}{0.8091} & 0.7325 & 0.7287 & 0.6857 & 0.5590 & 0.7059 & 0.7323 &\textcolor{blue}{\underline{0.7355}} \\
                                      & LPIPS~$\downarrow$  & \textcolor{red}{0.2134} & 0.2892 & 0.2364 & 0.2284 & 0.2177 & 0.2556 & 0.2328 & 0.4406 & \textcolor{blue}{\underline{0.2153}} \\
                                      & AHIQ~$\uparrow$     & 0.3746 & \textcolor{red}{0.4303} & 0.4047 & 0.3927 & 0.3841 & 0.3761 & \textcolor{blue}{\underline{0.4117}} & 0.3105 & 0.3985 \\
                                      & MUSIQ~$\uparrow$    & 64.73  & 55.57  & 67.42 & 64.65 & 67.33 & \textcolor{blue}{\underline{68.05}} & 66.73 & 67.84 & \textcolor{red}{68.81} \\
            \midrule
		\multirow{5}{*}{\makecell{RealSR \\ \cite{cai2019toward}}} & PSNR~$\uparrow$ & \textcolor{blue}{\underline{25.56}} & \textcolor{red}{25.87}  & 24.70 & 24.33  & 23.58 & 22.26 & 23.55 & 24.79 & 24.19 \\
		                          & SSIM~$\uparrow$  & 0.7390 & \textcolor{red}{0.7560}  & 0.7427 & 0.7456 & 0.7132 & 0.6462 & 0.7461 & 0.7423 & \textcolor{blue}{\underline{0.7485}}  \\
		                          & LPIPS~$\downarrow$  & 0.3719 & 0.3832 & 0.2713 & 0.2869 & 0.3016 & 0.3288 & \textcolor{red}{0.2543} & 0.2524 & \textcolor{blue}{\underline{0.2595}} \\
		                          & AHIQ~$\uparrow$  & 0.2370 & 0.2709 & \textcolor{red}{0.3167} & 0.3009 & 0.2965 & 0.2925 & \textcolor{blue}{\underline{0.3160}} & 0.3061 & 0.3145 \\
		                          & MUSIQ~$\uparrow$  & 41.69 & 35.19 & \textcolor{red}{64.62} & 57.55 & 60.53 & \textcolor{blue}{\underline{61.42}} & 59.92 & 59.67 & 60.47 \\
            \midrule
		\multirow{5}{*}{\makecell{DRealSR \\ \cite{wei2020component}}} & PSNR~$\uparrow$    & 27.79  & \textcolor{red}{27.96}  & 26.18 & 25.82  &24.56  & 23.39 & 24.85 & \textcolor{blue}{\underline{27.87}} & 25.67 \\
                                     & SSIM~$\uparrow$    & 0.8158 & \textcolor{blue}{\underline{0.8357}} & 0.7934  &0.7987  &0.7362  & 0.6723 & 0.8326 & 0.8056 & \textcolor{red}{0.8415} \\
                                     & LPIPS~$\downarrow$ & 0.3851 & 0.3933 & 0.2929  & \textcolor{blue}{\underline{0.2818}}  &0.3374  & 0.3860 & 0.2853 & 0.5408 & \textcolor{red}{0.2627} \\
                                     & AHIQ~$\uparrow$    & 0.2606 & 0.2887 & \textcolor{red}{0.3502}  &0.3478  &0.3192  & 0.2859 & 0.3489 & 0.2849 & \textcolor{blue}{\underline{0.3496}} \\
                                     & MUSIQ~$\uparrow$   & 22.41  & 21.88  & 35.50   & 35.25  &31.78  & 37.98 & 35.39 & \textcolor{red}{42.68} & \textcolor{blue}{\underline{38.66}} \\
            \midrule
		\multirow{4}{*}{\makecell{DPED-\\iPhone~\cite{ignatov2017dslr}}} & BRISQUE~$\downarrow$ & \textcolor{blue}{\underline{5.69}} & 46.95 & 15.89 & 16.49 & \textcolor{red}{3.46} & 22.12 & 13.31 & 13.43 & 11.19 \\
		                  & MUSIQ~$\uparrow$ & 51.27 & 39.81 & 51.65 & 50.99 & \textcolor{red}{57.19} & \textcolor{blue}{\underline{56.59}} & 49.97 & 46.87 & 51.84 \\
                            & TOPIQ~$\uparrow$ & 0.4675 & 0.3365 & 0.4879 & 0.4646 & \textcolor{red}{0.5439} & 0.4329 & 0.4437 & 0.4611 & \textcolor{blue}{\underline{0.4925}} \\
                            & NIQE~$\downarrow$ & 3.20 & 6.19 & 3.37 & \textcolor{blue}{\underline{3.17}} & 5.09 & 5.56 & 3.80 & 5.58 & \textcolor{red}{3.09} \\
        \bottomrule
	\end{tabular}}
\end{table*}

\subsection{Experimental Settings}
\label{subsec:expsetting}
\subsubsection{Datasets.} We follow the degradation pipeline of Real-ESRGAN~\cite{wang2021real} to synthesize LR-HR paired images on DIV2K~\cite{Agustsson_2017_CVPR_Workshops}, Flickr2K~\cite{timofte2017ntire}, and OutdoorSceneTraining datasets~\cite{wang2018recovering} for our training datasets. 
Then, we evaluate our method on both synthetic and real-world testsets. For synthetic data, we follow the above degradation strategy and generate LR-HR pairs from the DIV2K validation set~\cite{Agustsson_2017_CVPR_Workshops}. 
For real-world datasets, we follow the common settings to conduct comparisons on RealSR~\cite{cai2019toward}, DRealSR~\cite{wei2020component}, and DPED-iPhone~\cite{ignatov2017dslr} testsets.

\subsubsection{Evaluation Metrics}
For paired testsets~(\ie, DIV2K Valid~\cite{Agustsson_2017_CVPR_Workshops}, RealSR~\cite{cai2019toward}, and DRealSR~\cite{wei2020component}), follow previous works, we employ several perceptual-based metrics including LPIPS~\cite{zhang2018unreasonable}, AHIQ~\cite{lao2022attentions}, and MUSIQ~\cite{ke2021musiq} to measure the image perceptual quality. We also employ PSNR and SSIM metrics to indicate differences between different methods. 
For datasets without ground-truth image~(\ie, DPED-iphone~\cite{ignatov2017dslr}), follow previous works, we employ NR-IQA metrics including BRISQUE~\cite{Mittal2011blind}, TOPIQ~\cite{chen2023topiq}, NIQE~\cite{mittal2012making}, and MUSIQ~\cite{ke2021musiq} for perceptual quality.

\begin{figure}[h]
  \centering
   \includegraphics[width=\linewidth]{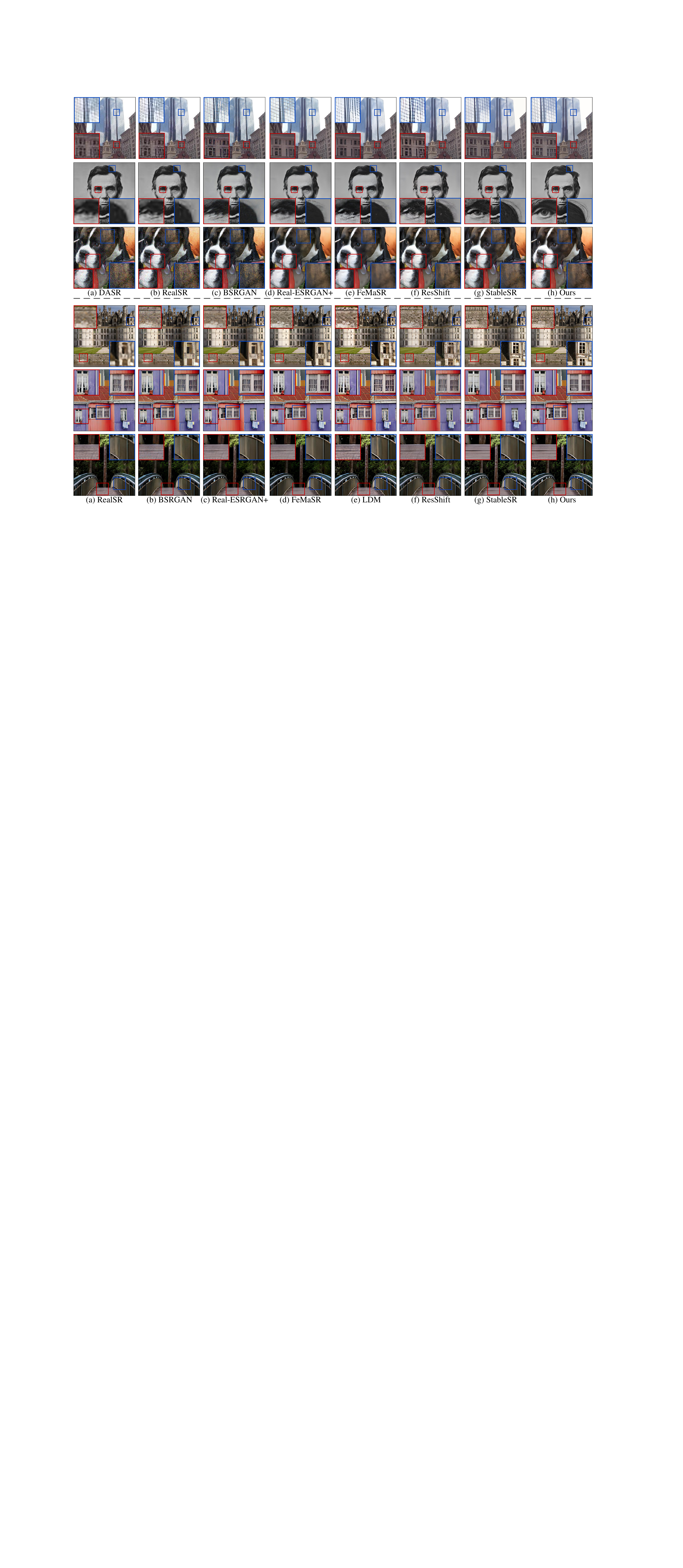}
   \caption{Qualitative comparisons on several real-world images. (The first three rows are arbitrary resolutions, and the resolution of the last three rows is fixed at $512 \times 512$)}
   \label{fig:bigpic}
\end{figure}

\subsection{Comparisons with Existing Methods}
\label{subsec:comparison}
To verify the effectiveness of our method, we compare AdaDiffSR with several state-of-the-art methods, including RealSR~\cite{ji2020real}, DASR~\cite{liang2022efficient}, BSRGAN~\cite{zhang2021blind}, Real-ESRGAN+~\cite{wang2021real}, FeMaSR~\cite{chen2022real}, LDM~\cite{rombach2022high}, StableSR~\cite{wang2023exploiting}, and ResShift~\cite{yue2024resshift}. For fair comparisons, we obtain the performance from the official code and models for testing. Note that the metrics reported for LDM~\cite{rombach2022high} in Tab.~\ref{tab:metrics} are obtained on the crop patches of corresponding datasets, because the LDM only supports static input resolution of $512 \times 512$.

\subsubsection{Quantitative Comparisons.} We first conduct the quantitative comparisons with several state-of-the-art methods on the synthetic dataset~\cite{timofte2017ntire} and three real-world benchmarks~\cite{cai2019toward, wei2020component, ignatov2017dslr}. As shown in Tab.~\ref{tab:metrics}, the proposed AdaDiffSR outperforms state-of-the-art methods on several perceptual-based metrics including LPIPS and NIQE. Specifically, on DRealSR, AdaDiffSR achieves a 0.2627 LPIPS score, which is 6.8\% better than Real-ESRGAN+ and 25.5\% better than other GAN-based SR approaches. Moreover, AdaDiffSR achieves comparable performance with other state-of-the-art methods in other benchmarks. 
Although RealSR and DASR achieve higher PSNR and SSIM metrics, they show relatively lower performance on perceptual-based metrics. As shown in Fig.~\ref{fig:bigpic}, they fail to reconstruct more realistic visual details, thus generating inferior visual results.

\begin{table}[h]\small
\setlength\tabcolsep{1pt}
\captionsetup{font=small}%
\begin{center}
\begin{minipage}{0.41\textwidth}
\caption{The complexity comparisons between DMs-based methods.}
\label{tab:complexity}
\centering
\resizebox{\linewidth}{!}{
\begin{tabular}{c|ccc}
		\toprule
            Methods & \#P(M) & FLOPs(G) & RT(s) \\ 
            \midrule
		\multirow{2}{*}{LDM~(2022)~\cite{rombach2022high}} & \multirow{2}{*}{113.60} & 888.9 & \textcolor{blue}{\underline{6.1}} \\ 
               &  & - & - \\
		\multirow{2}{*}{StableSR~(2023)~\cite{wang2023exploiting}} & \multirow{2}{*}{140.91}  & 1004.2 & 7.2 \\ 
                    &  & 4016.8 & \textcolor{blue}{\underline{25.9}} \\ 
            \multirow{2}{*}{ResShift~(2024)~\cite{yue2024resshift}} & \multirow{2}{*}{118.59}  & 922.3 & 19.7 \\ 
                    &  & 3689.2 & 71.3 \\ 
            \multirow{2}{*}{Ours} & \multirow{2}{*}{120.43} & 751.4 & \textcolor{red}{4.8} \\ 
                & & 2768.2 & \textcolor{red}{16.8} \\ 
        \bottomrule
	\end{tabular}}
\end{minipage}
\hfill
\begin{minipage}{0.55\textwidth}
\caption{Ablation studies of interval in skip-step on DRealSR dataset~\cite{wei2020component}.}
\label{tab:ablation_interval}
\centering
\resizebox{\linewidth}{!}{
\begin{tabular}{c|c|cccc}
        \toprule
	  Exp. & interval & PSNR~$\uparrow$ & SSIM~$\uparrow$ & LPIPS~$\downarrow$ & RT(s)\\
        \midrule
		(a) & (5,5,5,5) & 25.43 & \textcolor{red}{0.8437} & \textcolor{red}{0.2578} & 13.4\\
		(b) & (5,10,10,15) & 25.57 & \textcolor{blue}{\underline{0.8419}} & \textcolor{blue}{\underline{0.2607}} & 11.8 \\
		(c) & (5,10,15,25) & \textcolor{blue}{\underline{25.74}} & 0.8261 & 0.2754 & 8.9\\
		(d) & (5,15,20,25) & \textcolor{red}{25.82} & 0.8193 & 0.2796 & \textcolor{blue}{\underline{8.4}}\\
		(e) & (10,15,20,25)  & 24.83 & 0.7965 & 0.2939 & \textcolor{red}{7.3}\\
            Ours & (5,10,15,20) & 25.67 & 0.8415 & 0.2627 & 9.1\\
        \bottomrule
	\end{tabular}}
\end{minipage}
\end{center}
\end{table}

\subsubsection{Qualitative Comparisons.} To further compare the visual qualities of different approaches, we present visual results on real-world examples from real-world benchmarks and the internet in Fig.~\ref{fig:bigpic}. Due to the powerful generative capability of diffusion models, AdaDiffSR outperforms CNN-based and GAN-based methods in detail generation. In addition, it can be noticed that AdaDiffSR has significant improvements in realistic details compared to DMs-based methods, especially for surfaces and structures. For example, in the first and fourth rows in Fig.~\ref{fig:bigpic}, AdaDiffSR significantly enhances the texture details of building surfaces. More visual results can be found in the supplementary materials.

\begin{table}[h]
\setlength\tabcolsep{1pt}
\captionsetup{font=small}%
\begin{center}
\begin{minipage}{0.50\textwidth}
\caption{Ablation studies of IQA metrics used in MMLE on RealSR dataset.}
\label{tab:mmle_ablation}
\centering
\resizebox{\linewidth}{!}{
\begin{tabular}{c|cc|cccc}
        \toprule
	\multirow{2}{*}{Exp.} & \multicolumn{2}{c|}{Components} &   \multirow{2}{*}{PSNR~$\uparrow$} & \multirow{2}{*}{SSIM~$\uparrow$} & \multirow{2}{*}{LPIPS~$\downarrow$} & \multirow{2}{*}{MUSIQ~$\uparrow$} \\
        \cmidrule{2-3}
	& FR & NR &  &  &  &  \\
        \midrule
        (a)& \checkmark &  & \textcolor{red}{25.24} & \textcolor{red}{0.7532} &  0.3628  & 42.53 \\
	(b)& & \checkmark & 22.37 & 0.6872 & \textcolor{blue}{\underline{0.2714}} & \textcolor{blue}{\underline{51.23}} \\
	Ours& \checkmark & \checkmark & \textcolor{blue}{\underline{24.19}} & \textcolor{blue}{\underline{0.7485}} & \textcolor{red}{0.2595} & \textcolor{red}{60.47}  \\
        \bottomrule
\end{tabular}}
\end{minipage}
\hfill
\begin{minipage}{0.46\textwidth}
\caption{Ablation studies of PFJ on DPED-iPhone dataset~\cite{ignatov2017dslr}.}
\label{tab:ablation_pfj}
\centering
\resizebox{\linewidth}{!}{
\begin{tabular}{c|ccc|ccc}
        \toprule
	\multirow{2}{*}{Exp.} & \multicolumn{3}{c|}{Components} &   \multirow{2}{*}{MUSIQ~$\uparrow$} & \multirow{2}{*}{TOPIQ~$\uparrow$} & \multirow{2}{*}{NIQE~$\downarrow$}   \\
        \cline{2-4}
	& CC & CA & PFJ &  &  &  \\
        \midrule
        (a)& \checkmark &  &  & 32.19 &  0.3765  & 7.94 \\
	(b)& & \checkmark &  & \textcolor{blue}{\underline{42.37}} & \textcolor{blue}{\underline{0.4314}} & \textcolor{blue}{\underline{4.21}} \\
	Ours&  & & \checkmark & \textcolor{red}{51.84} & \textcolor{red}{0.4925} & \textcolor{red}{3.09}  \\ 
        \bottomrule
	\end{tabular}}
\end{minipage}
\end{center}
\end{table}

\subsubsection{Complexity Comparisons.} As for the complexity of AdaDiffSR, Tab.~\ref{tab:complexity} shows the detailed computational resources and inference time between different DMs-based SR methods. \#P and RT indicate parameters and runtime, respectively. 
For fair comparisons, we use the author-released models to report inference times and FLOPs for LR images with different resolutions running on the NVIDIA Tesla 32G-V100 GPUs. All approaches use the same equivalent timesteps as 50 in DDIM sampling strategy~\cite{song2020denoising} and input resolution as $512 \times 512$ or $1024 \times 1024$. 
Compared to the state-of-the-art SR approach StableSR, AdaDiffSR achieves comparable performance using $1.5 \times$ fewer inference times and $2.7 \times$ fewer FLOPs. 
We also find that adding the proposed approach to the existing DMs-based methods further improves the inference speed while maintaining reconstruction quality. More relative analysis can be found in the supplementary materials.

\subsection{Ablation Study}
\label{subsec:ablation}
In this section, we conduct the ablation on the different modules, analyze the visual superiority of progressive feature injection and region integration strategy.

\subsubsection{Effect of the interval in skip-step.} We first analyze the effectiveness of interval in skip-step. As mentioned in Sec.~\ref{subsec:dts}, we use different timesteps in different image regions, which is accomplished through skip-step strategy. 
We conduct several different interval settings to demonstrate the influence of intervals on DRealSR~\cite{wei2020component} dataset using DDPM sampling strategy. 
As shown in Tab.~\ref{tab:ablation_interval}, increasing the interval leads to a significant improvement in inference time. 
However, it also brings the degradation in overall visual performance and vice versa. Finally, we set the interval of skip-step as 5, 10, 15, and 20 to achieve a trade-off between utilization of computational resources and restoration quality.

\subsubsection{Effect of the metrics in MMLE.} 
\begin{figure}[t]
\centering
\begin{minipage}[t]{0.54\textwidth}
\centering
\includegraphics[width=\linewidth]{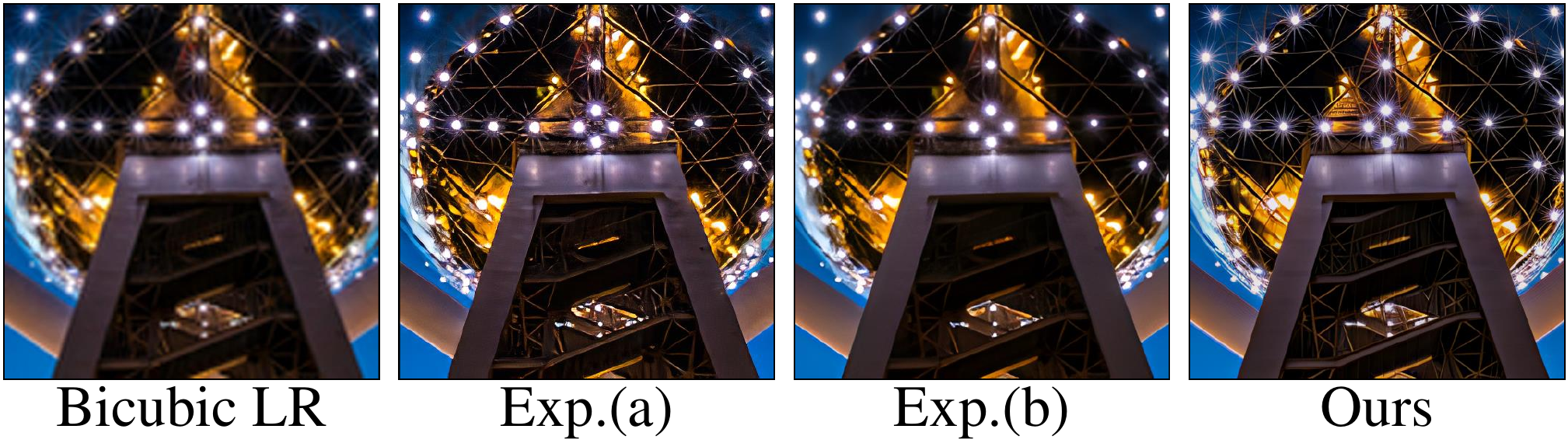}
\caption{Visual comparisons of different configurations in MMLE metrics.}
\label{fig:ablation_mmle}
\end{minipage}
\hfill
\begin{minipage}[t]{0.41\textwidth}
\centering
\includegraphics[width=1.0\linewidth]{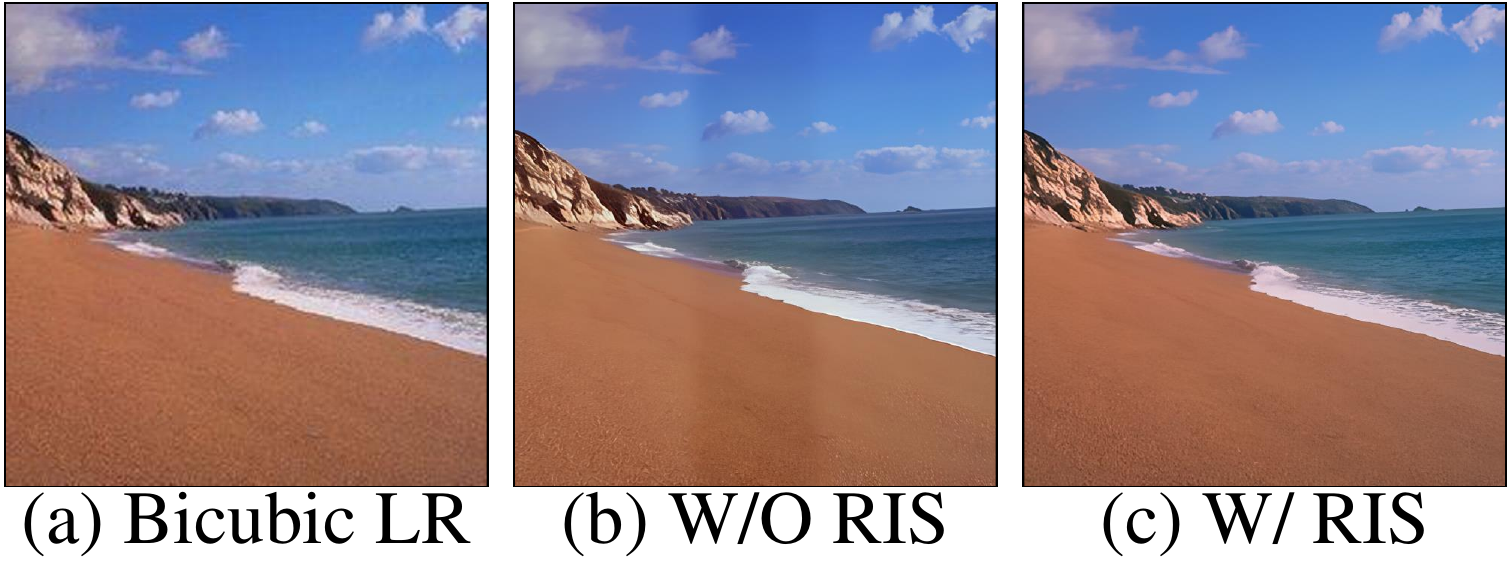}
\caption{Visualization results of using region integration strategy.}
\label{fig:integration}
\end{minipage}
\end{figure}

As mentioned in Sec.~\ref{subsec:mmle}, we choose four FR-IQA metrics and two NR-IQA metrics to measure the information gain during the denoising process. Here, we further verify the impact of the selected IQA metrics on RealSR~\cite{cai2019toward} dataset. As shown in Tab.~\ref{tab:mmle_ablation}, only FR-IQA metrics~(\ie, Exp.~(a)) or NR-IQA metrics~(\ie, Exp.~(b)) to guide the denoising process will lead to sub-optimal results. Specifically, only with FR metrics~(\eg, PSNR or SSIM) leads to the denoising process focusing on the restoration of latent features which consistent with the original image features, thus limiting the powerful generative capacity of the diffusion model, and vice versa. As shown in Fig.~\ref{fig:ablation_mmle}, we show visual comparisons to demonstrate the effectiveness of using both FR and NR metrics. More IQA-metrics analysis about mutual interaction within FR or NR metrics can be found in the supplementary materials.

\subsubsection{Effect of the progressive feature injection.}  To demonstrate the effectiveness of the PFJ module, we adopt several different configurations to validate the influence of the PFJ module. CC and CA indicate concat and cross-attention mechanisms, respectively.
As shown in Tab.~\ref{tab:ablation_pfj}, when original image features are fused in the concat mechanism, it leads to significant performance degradation. 
With the cross attention mechanism, the reconstruction performance improves obviously. 
Furthermore, the PFJ module uses modulation coefficients to regulate the original image features, and then injects these features into the denoising process, leading to a better trade-off between fidelity and realism.

\subsubsection{Effect between different region slicing strategy.} As mentioned in Sec.~\ref{subsec:tradeoff}, we slice the input LR images into multiple overlapping regions with the fixed resolution as the pre-trained DMs. However, the above slicing strategy is not consistent with the original intent of distinguishing between foreground and background regions as mentioned in Fig.~\ref{fig:teaser}. 
To demonstrate the effectiveness of the above slicing strategy, we apply super-pixel segmentation methods~\cite{kirillov2023segany, zhao2023fast, zhang2023faster, xiong2023efficientsam} to distinguish the foreground and background regions, and the experimental results are shown in Tab.~\ref{tab:ablation_patch}, the different slicing strategy has no significant impact on the visual results. However, due to the input resolution limitation of pre-trained diffusion models, we slice or fill these segmentation results into fixed resolution regions using zero-padding, then feed these regions into the proposed AdaDiffSR for inference, which may lead to the meaningless filled regions consuming massive computational resources, thus slowing the inference time. 
As shown in Fig.~\ref{fig:integration}, we visualize the effectiveness of the region integration strategy. With this strategy, the discontinuities disappeared significantly.

\begin{table}[t]\small
\setlength\tabcolsep{1pt}
\captionsetup{font=small}%
\begin{center}

\begin{minipage}{0.46\textwidth}
\caption{Ablation studies of region slicing strategy on DIV2K Valid dataset~\cite{timofte2017ntire}. We report the inference time on the $512\times 512$ resolution.}
\label{tab:ablation_patch}
\centering
\resizebox{\linewidth}{!}{
\begin{tabular}{c|ccccc}
        \toprule
        
	  Exp. & \#P(M) & PSNR$\uparrow$ & SSIM$\uparrow$ & LPIPS$\downarrow$ & RT(s)\\
        
        \midrule
	    SAM~\cite{kirillov2023segany} & 636 & 24.21 & 0.7368 & 0.2232 & 9.43\\
		FastSAM~\cite{zhao2023fast} & 68 & 24.11 & 0.7323 & 0.2182 & 8.79\\
		MobileSAM~\cite{zhang2023faster} & 10 & 24.14 & 0.7363 & 0.2178 & \textcolor{blue}{\underline{8.59}}\\
		EfficientSAM~\cite{xiong2023efficientsam} & 25 & 24.36 & 0.7311 & 0.2204 & 8.72\\
            Ours & 0 & 24.25 & 0.7355 & 0.2153 & \textcolor{red}{4.81}\\
        \bottomrule
	\end{tabular}}
\end{minipage}
\hfill
\begin{minipage}{0.50\textwidth}
\caption{Ablation studies of different modules in AdaDiffSR on DIV2K Valid dataset~\cite{timofte2017ntire}. We report the inference time on the $1024\times 1024$ resolution.}
\label{tab:ablation_modules}
\centering
\resizebox{\linewidth}{!}{
\begin{tabular}{cc|ccccc}
        \toprule

	  Exp. & Models. & \#P(M) & PSNR$\uparrow$ & SSIM$\uparrow$ & LPIPS$\downarrow$ & RT(s)\\
        
        \midrule	
        Baseline &    LDM~\cite{rombach2022high} & 113.60 & 20.58 & 0.5590 & 0.2556 & 22.3\\
	\multirow{2}{*}{(a)} &	+ DTSS~(IQA) & 113.60 & 22.16 & 0.5377 & 0.2843 & 21.5\\
	&	+ DTSS~(MMLE) & 115.79 & 22.38 & 0.5294 & 0.2715 & 18.1\\
	(b)&	+ PFJ & 120.43 & \textcolor{blue}{\underline{24.22}} & \textcolor{blue}{\underline{0.7299}} & \textcolor{blue}{\underline{0.2338}} & \textcolor{red}{16.5}\\
        (c)&    + RIS~(Ours) & 120.43 & \textcolor{red}{24.25} & \textcolor{red}{0.7355} & \textcolor{red}{0.2153} & \textcolor{blue}{\underline{16.8}}\\
        \bottomrule
	\end{tabular}}
\end{minipage}
\end{center}
\end{table}

\subsubsection{Effect of different modules in AdaDiffSR.} To demonstrate the effectiveness of components in AdaDiffSR, we add each component to the baseline model LDM, and the results are shown in Tab.~\ref{tab:ablation_modules}. DTSS~(IQA) and DTSS~(MMLE) denote that calculate the IQA metrics using PYIQA toolbox~\cite{pyiqa}~(\ie, the ground-truth) and using regressor, respectively. There is no significant gap between the above strategy, but the MMLE regressor would reduce the inference time considerably. Besides, with the addition of the PFJ module, the experimental results and inference time become better. Although the region integration strategy~(RIS) did not contribute significantly to the metrics, as shown in Fig.~\ref{fig:integration}, this strategy eliminated the discontinuity, thus improving the visual quality.

\section{Limitations}
\label{sec:limit}
Although the proposed AdaDiffSR achieves the trade-off between computational resources and restoration quality compared to other DMs-based SR methods. However, the DMs-based SR approaches require multiple denoising steps to obtain the restored results, which is slower than previous CNN or GAN-based SR methods. 
Moreover, since the fixed input resolution limitations of the pre-trained DMs, the proposed method applies the static slicing strategy, which is somewhat different from the original design intention mentioned in Fig.~\ref{fig:teaser}. 
Our future works will concentrate on investigating new mechanisms that acclerate the denoising process, and achieving more fine-grained region reconstruction.

\section{Conclusion}
\label{sec:conclusion}
In this paper, we explore the trends that apply adaptive inference strategy on diffusion models based SR methods. 
To this end, we propose a novel DMs-based SR pipeline AdaDiffSR for real-world image super-resolution, which utilizes the latent information gain to adjust timesteps dynamically during the denoising process and effectively reduces the computational resource overhead. 
Experimental results on both synthetic and real-world benchmarks demonstrate that the proposed AdaDiffSR achieves a better trade-off between computational resources and restoration quality. We believe that our exploration will provide more inspiration for future works.

\section*{Acknowledgements}
This work was supported in part by the NSFC under Grant 62272380 and 62103317, the Fundamental Research Funds for the Central Universities, China (xzy022023051), the Innovative Leading Talents Scholarship of Xi'an Jiaotong University.

%


\end{document}